\documentclass[letterpaper]{article}
\usepackage{aaai}
\usepackage{times}
\usepackage{helvet}
\usepackage{courier}
\usepackage{graphicx}
\author{Isaac Kamlish \\ 
 University College London \\
 \texttt{zcahika@ucl.ac.uk} \\\And
 Isaac Bentata Chocron \\ 
 University College London \\
 \texttt{zcecibe@ucl.ac.uk} \\\And
 Nicholas McCarthy \\ 
 University College London \\
 \texttt{zcapnmc@ucl.ac.uk} \\
}

\begin{document}
\title{SentiMATE: Learning to play Chess through Natural Language Processing}
\maketitle
\begin{abstract}
 We present SentiMATE, a novel end-to-end Deep Learning model for Chess, employing Natural Language Processing that aims to learn an effective evaluation function assessing move quality. This function is pre-trained on the sentiment of commentary associated with the training moves, and is used to guide and optimize the agent's game-playing decision making. The contributions of this research are three-fold: we build and put forward both a classifier which extracts commentary describing the quality of Chess moves in vast commentary datasets, and a Sentiment Analysis model trained on Chess commentary to accurately predict the quality of said moves, to then use those predictions to evaluate the optimal next move of a Chess agent. Both classifiers achieve over 90\% classification accuracy. Lastly, we present a Chess engine, SentiMATE, which evaluates Chess moves based on a pre-trained sentiment evaluation function. Our results exhibit strong evidence to support our initial hypothesis -  \textit{``Can Natural Language Processing be used to train a novel and sample efficient evaluation function in Chess Engines?''} - as we integrate our evaluation function into modern Chess engines and play against agents with traditional Chess move evaluation functions, beating both random agents and a DeepChess implementation at a level-one search depth - representing the number of moves a traditional Chess agent (employing the alpha-beta search algorithm) looks ahead in order to evaluate a given chess state. 
\end{abstract}
\section{Introduction}
Chess has traditionally been one of the most challenging and researched Artificial Intelligence problems. Modern Chess computational theory is reliant on two key components: a search algorithm that builds a game tree by exploring all possible moves, and an evaluation algorithm, which analyses the advantage of moving to a given node within the game tree \cite{Heath}. Our contribution to this area focuses on developing a novel evaluation method of a Chess move, based on using moves which have been categorised as `positive' given the sentiment of the commentary describing it, using a Sentiment Analysis model.
\\
\\
Building algorithms with superhuman ability to play Chess has been accomplished by systems such as IBM's DeepBlue. However, machine learning alternatives, such as DeepMind's AlphaZero, only succeeded after millions of iterations of self-play and using thousands of Tensor Processing Units (TPUs) - not generally available to the research community \cite{Silveretal}. There is an estimated $10^{120}$ possible Chess game states ~\cite{Shannon}, and $10^{47}$ reachable positions \cite{Chinchalkar} on a Chess board. Thus, the ability of an algorithm to outperform State of the art (SOTA) is reliant on vast amounts of computational power to model the depth of the aforementioned game tree, and the ability of the evaluation function to select the optimal action. Current Deep Learning approaches to Chess use neural networks to learn their evaluation function. Supervised attempts at this have created their training set by sampling states at random from professional Chess games which resulted in victory ~\cite{David}. The sampling from victorious games results in the loss of individual information regarding the sampled move, and reveals little information about the move itself. Additionally, early Chess states in winning games are not necessarily 'winning' states. Hence, training such a model may be inefficient, and will require massive amounts of samples to allow the model to generalize well.
\\
\\
In our research, we aim to tackle the assessment of the quality of individual movements through the use of Natural Language Processing, to avoid this loss of information before evaluating the optimality of each node within the tree. Data from different Chess websites was scraped, which included information regarding the moves being made, and a qualitative assessment of the moves themselves in the form of commentary, written by a wide range of Chess players; resulting in a large database of moves with annotated commentary.
\\
\\
The data was cleaned so that only commentaries which were deemed to describe the `quality' of the moves were employed, through the use of a preliminary `Quality vs Non-Quality' classifier, trained using 2,700 commentaries which we hand-labelled. Hand-engineering was then applied to these results to ensure False-Positives for the `Quality' section were removed, ultimately resulting in a dataset of `Quality' moves on which Sentiment Analysis was to be performed. A novel Sentiment Analysis model was built, given that the nature of Chess commentary is different to that of, for example, movie reviews, thus making sentiment classifiers such as that trained on the IMDb's database unsuitable for this problem. 2,090 moves were annotated for their `sentiment', and a bi-directional LSTM using stacked BERT and GloVe embeddings was trained on this commentary. The resulting dataset was then fed into our novel evaluation function for Chess moves trained on sentiment, which is tested against other Chess engines.
\newpage
\section{Related Work}
Historically, State Of The Art Chess engines have been designed using extensive manual feature selection to guide action/move selection in Chess games along with large amounts of specific domain knowledge - something we wanted to avoid. More recently however, high performance, alternative Chess engines such as Giraffe \cite{Giraffe} (a Reinforcement Learning, self-playing Chess engine) and DeepChess offered the first end-to-end machine learning-based models, using both unsupervised pretraining and supervised training, to exhibit Grand Master-level Chess-playing performance. \cite{David}. Seminal work done by Sabatelli \cite{Sabatelli}  went further and provided a way for Artificial Neural Networks to play high-level Chess without having to rely on any of the lookahead algorithms that are used by the above algorithms. Motivated by the success of this model to learn the quality of a Chess board state from an algebraic, or in our case, bit format input, we set out to use this Deep Learning architecture as the basis for our own evaluation function. To this end, we aim to employ a novel Sentiment Analysis model with a different focus: to influence and improve the decision making of a novel Chess playing agent. In this sense, the prediction of sentiment is not the end goal of the model and acts as a view into the inner workings of the Chess agent. Given the modular nature of our model, it provided us with an opportunity to perform important ablation studies in order to iteratively adapt and improve the various parts of the model so as to improve game performance.
\section{Background}
\begin{figure*}
  \centering
  \includegraphics[width=16cm]{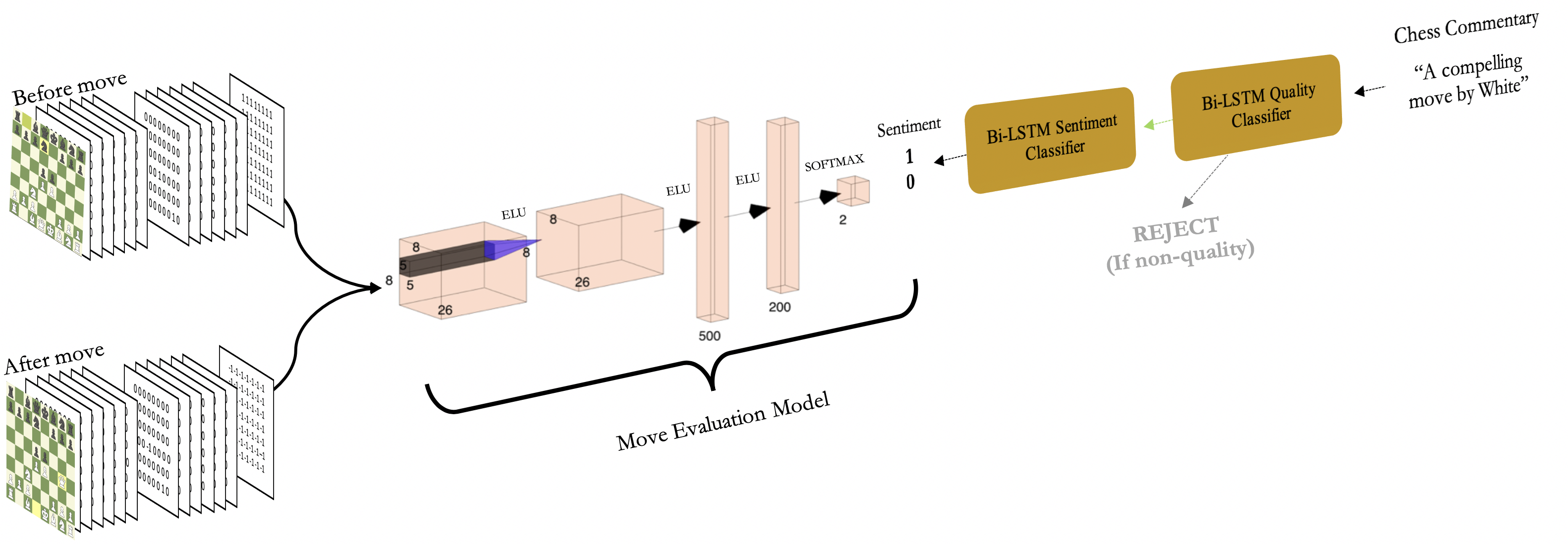}
  \caption{Complete pipeline for training the evaluation model} \label{biltong}
\end{figure*}

From the turn of the century, Sentiment Analysis has been one of the most active research areas in NLP, thus providing us with a platform from which to work on. To this end, we aim to employ SOTA Sentiment Analysis so as to influence and improve the decision making of a Chess playing agent. More recently, given the rapid gains in compute power available, Deep Learning has emerged as the dominant machine learning technique for Sentiment Analysis. Given the sequential nature of textual data, the effective use of RNNs (and in particular LSTM networks) has grown dramatically~\cite{zhang}, with the network being able to learn long-term dependencies between words. Given the nature of the text we are dealing with - short, emotive descriptions of a Chess move in online forums - we are focused on researching and utilising sentence level sentiment classification. Going beyond using a single neural network for sentence level sentiment classification, work has been done combining a CNN and RNN architecture \cite{Wang2} in order to exploit the coarse-grained local features generated by the CNN and the long-distance dependencies learned via the RNN. As was discovered, this required significantly more data preprocessing and data extraction than was originally established. 
\\
\\
Before Deep Learning led the way in Sentiment Analysis in NLP, substantial amount of work done had been done in applying Artificial Intelligence to the game of Chess. The ideal Chess agent does not only need the ability to evaluate each game state (the arrangement of pieces on the board) but also needs to be able to efficiently search through every potential move available to it - the exact role of the tree searching algorithm present in previous SOTA Chess engines \cite{Heath}. With infinite time and processing power, one would ideally explore all possible states, however, these constraints have lead to innovation in search algorithms to reduce the number of moves which need to be explored.
\\
\\
As mentioned, the second fundamental part to any Chess engine is the ability to accurately evaluate the quality of a configuration of pieces for a given player, which takes the form of a probability for each player winning from that given state. Furthermore, given the difficulties in tuning large numbers of parameters by hand, the number of features in previous evaluation functions is not that large. Deep Learning can help to alleviate these issues; not only can it learn all the necessary parameters, but due to the non-linearity of certain layers and large amount of parameters, Deep Learning Networks can allow our evaluation functions to be far more expressive than previous traditional linear formulas. \textit{DeepChess} receives 2 different boards as input, and utilises as siamese network to output a score on which one is better. \textit{Giraffe} \cite{Giraffe} initially trains its neural network for its evaluation function to predict the output of \textit{StockFish} scoring function by training it on 5 million states uniformly sampled from databases from either human Grand Masters or computers with its corresponding \textit{StockFish} score. The whole model was then trained on a dataset consisting of a hybrid of self-play and professional Chess-engine games \cite{Hassabis}.
\\
\\
All of these approaches to Chess tackled one fundamental issue: what data should be used to train these evaluation function neural networks? Deep Learning approaches such as \textit{DeepChess} labelled states depending on the games final outcome such that states were scored +1 if the game ended up in  a win, 0 for a draw and -1 for a loss. Even though the information is objective in the sense that a game was won/drawn/lost from that given state, an early state could be unrelated to the final outcome.  We believe using NLP move-based evaluation will improve this, ultimately reducing sample complexity. Figure \ref{biltong} shows the proposed pipeline for the model.
\section{Methods}
\textbf{Collection and analysis of data}
\\
The initial dataset comprised of approximately 300,000 individually commented Chess moves taken from a  online Chess discussion forum \footnote{https://gameknot.com/} using \cite{jhamtani2018learning}'s data scraper . The commentary was split up into five different categories, based on what it describes: `Move Quality', `Comparative', `Planning', `Contextual Information', `General', using the provided pre-trained SVM classifier. The scope of this project draws upon applying Sentiment Analysis to `Quality Moves', thus making it the most significant category of our analysis. Upon inspection of the provided data's accuracy, it was found that, whilst `Quality' commentary was categorised with 90\% accuracy, the accuracy of commentary classification for the remainder commentary hovered around 70\%, thus leading to a significant amount of false positives being categorised as `Quality', resulting in a dataset riddled with Non-Quality moves. Furthermore, the boundaries applied to the `Quality' moves were too strict, resulting on classification of `Quality' moves into other categories given this was how the classifier was trained. This resulted in a dataset of only about 1,000 commentaries deemed to be describing the true `Quality' of moves, insufficient to train both a sentiment model and a Chess engine.
\\
\\
To remediate this, we propose a newly trained classifier based on a more relaxed `Quality' commentary boundary and binary classification (Quality VS Non-Quality), to ensure better `Quality' classification, and a lower quantity of False Positives. To this end, we manually labelled 2,700 Chess commentaries for training and 300 commentaries for testing, in the binary class setting. Alternative data avenues were also pursued, expanding the dataset using information from six additional websites.  Furthermore, the data was refined to ensure better classification quality: all moves represented by a single token (e.g., a move commentary that is stated in Chess notation - Bxg8, Kf8, Nf3), commentaries suffixed with seemingly arbitrary numbers (e.g., ``What a fantastic move 95"), and punctuation that does not pertain to standardised notation (and that would not contribute to any semantic understanding of the text), were removed from the data.
\\
\\
Inspired by Zhang et al.'s work on RNNs and LSTMs for text classification ~\cite{zhang}, we set out to experiment with different structures of Recurrent Neural Networks, alongside different word embeddings, in order to find those yielding the highest accuracy of classification into a binary setting using the previously produced training set: we consider a vanilla RNN, an LSTM, and a bi-directional LSTM, with different permutations of stacked and non-stacked GloVe and BERT embeddings, as potential models for text classification. The results and discussion can be found on `Experiments' and `Results \& Discussion' respectively.
\\
\\
With a condensed, structured and quality dataset containing both the text-based commentary and state representation (before and after the move has occurred), we proceed to format the state representations into our novel move representation which eventually can be feed directly into the evaluation function, as seen in Figure 1. The provided, standardised format for the state representations  is given in Portable Game Notation (PGN) format - a plain text computer-processible format for recording Chess games.  The state representations for before and after the given move are then converted to FEN format using Python-chess    \footnote{https://github.com/niklasf/python-Chess}). These FEN formats are then used to extract the final representation.
\\
\\
The final representation for each of the 2 states (before and after the move) are transformed into a $8 \times 8 \times 12$ representation of the Chess board - where the first two dimensions corresponding to the size of the Chess board and each of the 12 channels in the final dimension represent a different piece. $+1$'s are placed at points where white pieces are found, $-1$'s for blacks and $0$'s at points where no pieces are found. (i.e if there is a white queen at E4, a $+1$ would be placed at the co-ordinate (5,4) on the channel which represents queens). We then introduce a novel channel ($12 \rightarrow 13$) to signify whether it is black or white to move - (all $+1$'s for white pieces, all $-1$'s for black pieces). Finally, we stack the two vectors (pre-move and post-move) on top of each other to gain a final representation of $8 \times 8 \times 26$. This is the final data representation that we will feed into our CNN evaluation function.This work ultimately aims to construct a model that takes in raw natural language in the form of Chess-move commentaries, performs accurate classification into quality and non-quality, transitions through two different Chess-game data formats and outputs a $8 \times 8 \times 26$ representation of every Chess board state. The next key stage in our methodology focuses on taking this newly generated dataset and generating SOTA sentence level sentiment predictions using Deep Learning techniques.
\\
\\
\noindent \textbf{Sentiment Analysis}
\\
Given the ambiguous and neutral nature of some of the raw data Chess commentaries, performing Sentiment Analysis on the different categories would prove to be a challenging endeavour. For example, protecting a piece can be positive in that it avoids a negative consequence, or negative in that it shows the state of threat in the current game state. Furthermore, on pre-trained classifiers, ``An attack on the queen'' would generally be classified with negative sentiment given the nature of the word ``attack'' - however, this move would in reality be beneficial for the attacking player. Semantics and the contextual information are a key element to being able to correctly analyse the given data ~\cite{tetali}.
\\
\\
The first model we used for sentiment prediction was the jointed CNN-RNN architecture, trained on the Stanford Sentiment Treebank using word2vec embeddings pretrained on Google News. However, this model was seen to underperform the architecture put forward by Zalando's Flair in preliminary tests, thus making the latter our chosen architecture for research around Sentiment Analysis. Since the dataset commentaries are short pieces of text that often lack of features for effective classification (particularly fine-grained classification), the combination of an LSTM and appropriate word embeddings are extremely effective in learning word usage in the context of Chess games, given enough training data.  While we hypothesised, given the semantically narrow nature of our dataset (i.e discussing only Chess moves), that word embeddings such as word2vec would provide satisfactory results, we believed that the use of additional \textit{contextual string embeddings} would provide, in theory, a significant improvement in accuracy. These word embeddings capture latent syntactic-semantic information that goes beyond standard word embeddings, with two key differences: 1. they are trained without any explicit notion of words and 2. they are \textit{contextualised} by their surrounding text - meaning that same word will have different embeddings dependent on its contextual use. This latter point is extremely important in accurately predicting samples that are ambiguous in their sentiment and for a training set that is relatively small (compared with other Deep Learning training datasets).
\\
\\
\textbf{Move Evaluation}
\\
With a dataset containing the representations for the Chess board before and after moves - the $8 \times 8 \times 26$ tensor -  along with the sentiment for an array of Chess moves we then proceeded to input this into a self-built Convolutional Neural Network. After experimenting with a variety of models, the network architecture that we employed was based on work done by Sabatelli \cite{Sabatelli} - however, we propose a novel representation  that inputs a tensor of size $8 \times 8 \times 26$, applies two layers of convolutions (the first with filter size 5, the second with filter size 3) with \textit{elu} activations (the use of \textit{elu} can increase on the speed of convergence and gain a better accuracy \cite{elu}), has a dropout of $0.25$, is flattened and then followed by two dense, fully connected layers of size $500$ and $200$ respectively (both with \textit{elu} activations) to then finally output a binary, softmax output which quantifies the `goodness' ($G$) or `badness' ($B)$ of a move $M$, where $G, B \in [0,1]$. Given the importance of the location of every Chess piece on the board, padding techniques were employed while pooling were not, so as to preserve all the geometrical properties of the input - this way the output of the convolutions is of the exact same size as the original Chess positions.  The `goodness' output of each legal move $G(M)$ at time $t$ is recursively compared to other legal moves through the following proposed move search algorithm:
\\
\\
\textbf{Alpha Beta Move Search (ABMS)}
\\
To develop an algorithm which now searches between potential moves rather than states, we applied a slight tweak to the traditionally used Alpha-Beta search algorithm. Nodes still represent states $S_t$ however as we traverse down the tree we also store a variable which maintains the prior board state to the current node we are in $S_{t-1}$. We then can concatenate these two board states together to give us the input $ I = [R(S_{t-1}),R(S_t)]$  for our evaluation function. Given the Branching factor B and search Depth D, this search algorithm still maintains the reduced search complexity of $B^{\frac{D}{2}}$ while allowing us to evaluate in a move-wise fashion.
\section{Experiments and Results}
\textbf{Quality move category classification}
\\
Using the aforementioned training set of 2,700 labelled commentaries into quality and non-quality data, we proceed to experiment using different architectures for text classification with pre-trained GloVe embeddings. The considered architectures are: a vanilla RNN, an LSTM and a bi-directional LSTM. The parameters used in these architectures were fine-tuned using a hyper-parameter search, and yielded the following results on the testing set:\\
\begin{figure}[h]
  \centering
  \includegraphics[width=240pt]{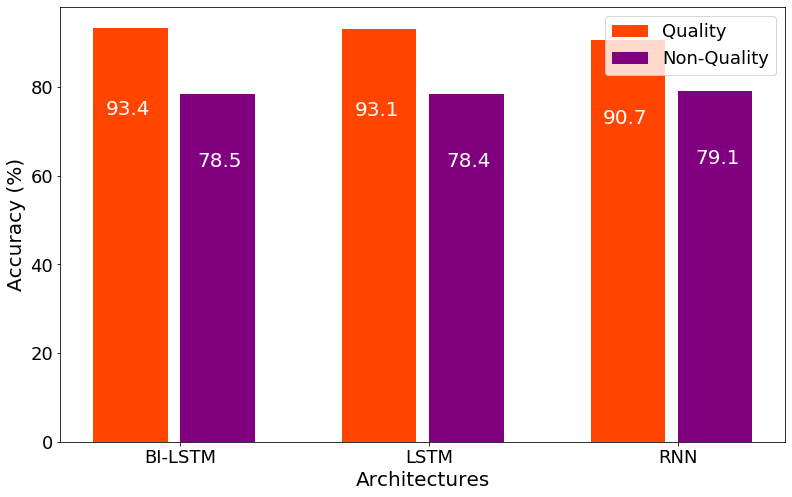}
  \caption{Analysis of different architectures for text classification} 
\end{figure}
\\
\noindent It can be inferred from Figure 2 that a bi-directional LSTM yields the highest classification results, with 94.2\% accuracy on classifying Quality data (0.8\% higher than the second-highest, the non bi-directional LSTM), and only marginally lower classification accuracy than the highest classifier for the non-Quality data (78.8\%, 0.2\% less than the non bi-directional LSTM). Upon this result, different word embeddings were tested using the bi-directional LSTM architecture, namely BERT, GloVe and stacked BERT and GloVe embeddings, the results of which can be seen in Figure 3.
\\
\begin{figure}[ht]
  \centering
  \includegraphics[width=240pt]{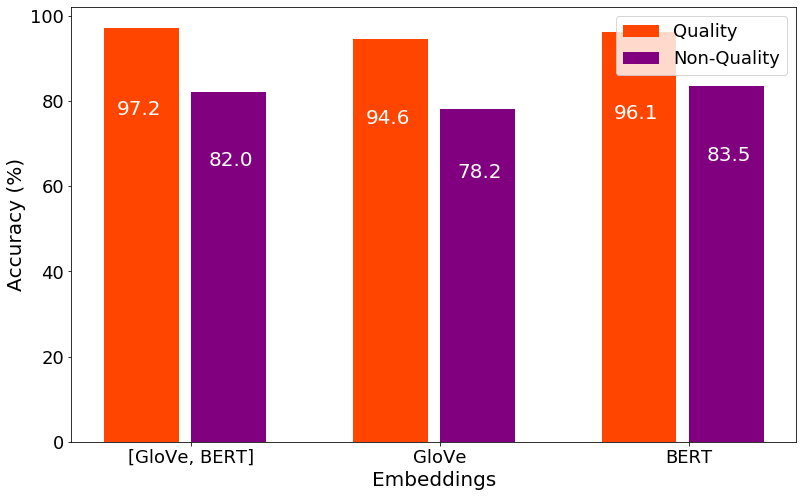}
  \caption{Analysis of different embeddings for text classification} 
\end{figure}
\\ 
\noindent BERT word embeddings achieved marginally higher results than the stacked GloVe and BERT embeddings, with 96.3\% classification accuracy for `Quality' commentary, and 84\% accuracy for non-Quality commentary respectively. This exceeds the classification accuracy reported in literature using SVMs, by 5\% for `Quality' commentary, and 9\% for non-Quality commentary (across the five aforementioned categories).
\\
\\
\noindent \textbf{Sentiment Classification}
\\
The LSTM sentiment model was trained on 2090 examples of moves that are considered to be informative about whether a move is good or bad (i.e \textit{a quality move description}). It was then tested on $233$ Chess commentaries that consisted only of \textit{quality moves} (commentary that describes whether the move is good or not), resulting in an accuracy of 91.42\%  and a balanced accuracy \cite{brodersen2010balanced} of 90.83\% - similar to that of the jointed CNN-RNN architecture (89.95\%) and other implementations of a bi-LSTM\cite{Barnes:journals/corr/abs-1709-04219} (82.6\%) in the binary sentiment prediction setting (when trained on the Stanford Sentiment Treebank dataset \footnote{https://nlp.stanford.edu/sentiment/}).
\\
\\
\textbf{Overall Chess agent performance}
\\
The final experiment involved the testing of the agent in its entirety, by playing it against an \textit{DeepChess} implemnentation and an agent with a random strategy. To be able to test our engine against a competitor over many games, the opponent's strategy must contain some element of stochastity. Otherwise the same sequence of moves will be played out for every game. To evaluate the performance of the agents, we used StockFish's evaluation function to evaluate how well SentiMATE was performing throughout the game. Due to the computational constraints on the look-ahead element of the agent, it struggles during the end game to checkmate its opponent, therefore a win was decided by the player with the best state (evaluated by StockFish) after 40 moves. We also measured the Material score for each player at every move using the most common assignment of points for each piece (Q:9, R:5, N:3, B:3, P:1).  Here the material score is defined to be the sum of the piece values (a value determining the relative strength of pieces in potential exchanges)  of each side. In 100 games, SentiMATE won 81\% of the time against the random agent and beat DeepChess on search depth 1 (representing the number of moves ahead the agent will look in order to evaluate it's current game state) as both black and white (given there is no randomness in both agents' play, only two games are possible). \\
\begin{figure}[h]
    \centering
    \includegraphics[width=220pt]{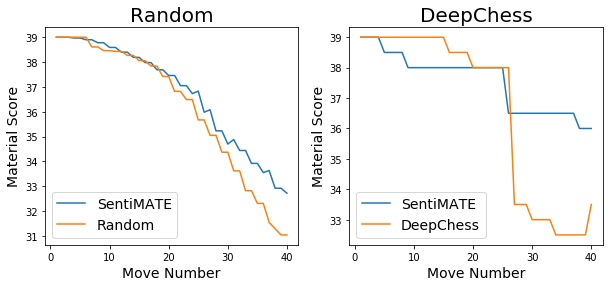}
    \caption{Material Score decay over time averaged over 100 games for Random and Black and White for DeepChess} 
\end{figure}
\noindent A positive divergence in material score can be seen in Figure 4 between SentiMATE and the Random agent, with the gap increasing as the game goes on.  This is suggestive that SentiMATE's strength occurs later in midgame and shows no strategical advantage during opening play. This issue has been noted by many previous Chess engines which typically use an opening book to guide the agent into middle play. A similar pattern can be seen against DeepChess where SentiMATE's advantage occurs after some exchanges after move 25. 
\section{Discussion}
\textbf{Move Quality}
\\
As mentioned in the `Experiments' section, the best performing model for `Quality' commentary classification was a bi-directional LSTM architecture using pre-trained BERT embeddings. These results were in accordance with our expectations: the LSTM exhibits SOTA performance, however the bi-LSTM is superior in this setting \cite{ZHOU:journals/corr/ZhouQZXBX16} - given the ability of the network to preserve information from both past and future (rather than just past information) thus understanding the context of each word in the commentary and returning accurate text classification. Upon classification of the provided dataset employing the aforementioned model, further tweaking was applied to the dataset. For example, the classifier was not trained on comments which reflected quality based solely on symbols common to Chess notation (e.g., !!, representing a good move), since previous experiments reflected that the use of these symbols confused the classifier in understanding the sentence structure and common words behind `Quality Moves'. Hence, these comments were added into the dataset. Upon cleaning and classifying the dataset based on commentary, bitifying the Chess moves, and applying Sentiment Analysis to the commentary, we present SentiChess
a dataset of 15,000 Chess moves represented in bit format, alongside their commentary and sentiment evaluation. This dataset is offered in the hope of further development of work around sentiment-based Chess models, and statistical move analysis.
\\
\\
\textbf{Sentiment prediction}
\\
Due to the required data format required by the jointed CNN-RNN architecture that we intended to use, we changed track and built our own bi-LSTM model to predict the sentiment of the Chess commentaries using stacked, pre-trained BERT and GloVe word embeddings in order to achieve the best results.\\ \\
While we were aware that fine-grained sentiment classification historically has displayed mediocre results, we postulated that a binary prediction would simply not provide enough information at the evaluation stage. Despite this, our prediction accuracies (and corresponding precision and recall) emerged unsatisfactory in the fine-grained setting using our custom built bi-LSTM. To this end, we revised the model to output binary predictions (good or bad move) with corresponding probabilities. It was found, after experimenting with various pre-trained word embeddings, that BERT embeddings (an embedding based on a bidirectional transformer architecture) were the most effective in accurate sentiment predictions. The particular embedding that enabled SOTA prediction consisted of 24 layers, 1024 hidden units, 16-heads and 340m parameters.  While character embeddings were tried and tested, they did not provide optimal test accuracy.  In the below, 1 (1.0) = Bad move with total certainty and 2 (0.05) = Good move with little certainty. The model displayed high levels of understanding of the commentaries, correctly predicting moves constructed with a variety of double negation techniques (implicit negation, negational prefixes and unseen double negatives). There were mixed results however when faced with unseen, emotive and descriptive adjectives and seemed to break down rather dramatically when faced with increasingly rare adjectives unseen in the training set, as can be observed in Figure 5:
\begin{figure}[h]
  \centering
  \includegraphics[width=210pt]{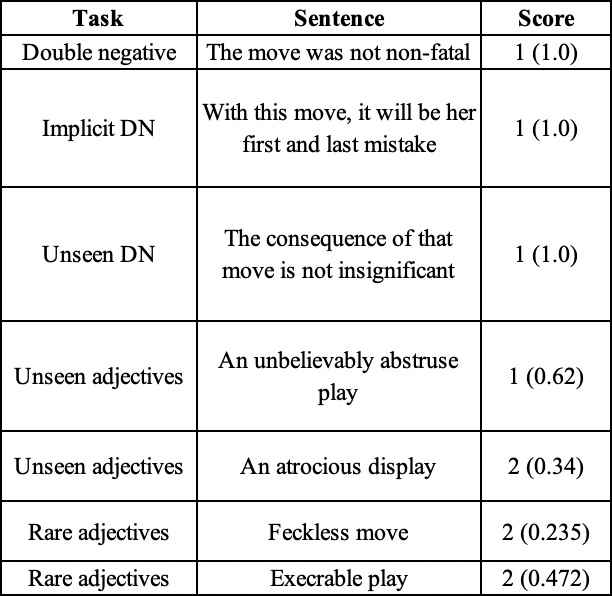}
  \caption{Sentiment Prediction on difficult sentences }
\end{figure}


\noindent Another important testing strategy was to establish how well the model dealt with annotation symbols from Algebraic notation \footnote{https://en.wikipedia.org/wiki/Algebraic\_notation\_(Chess)} - specifically the following symbols: \textbf{!:} an excellent move; \textbf{!!:} a particularly good move;  \textbf{?:} a bad move; a mistake; \textbf{??:} a blunder;  \textbf{!?:} interesting move that may not be best; \textbf{?!:} a dubious move or move that may turn out to be bad.\\ \\
It was shown that the model predicted correctly with total certainty on all testing examples that contained particularly ambiguous commentary descriptions of abnormally large lengths ($\approx$100 words), but were prepended with one of the above symbols (or permutations thereof).
\\
\\
\textbf{Chess Model}
\\
To delve deeper into the strengths and weaknesses of the model, the positions of each of SentiMATE's pieces over the 100 games were logged. There is a clear distinction between the pieces which the agent has managed to deploy successfully and ones which it has not. The heatmap in Figure 6 display the development of pieces by SentiMATE:
\begin{figure}[h]
  \centering
  \includegraphics[width=230pt]{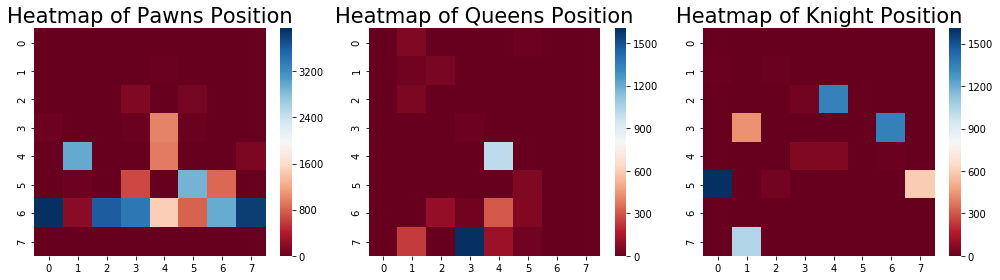}
  \caption{Heatmaps of SentiMATE piece's which show positive development}
\end{figure}

\begin{figure}[h]
  \centering
  \includegraphics[width=230pt]{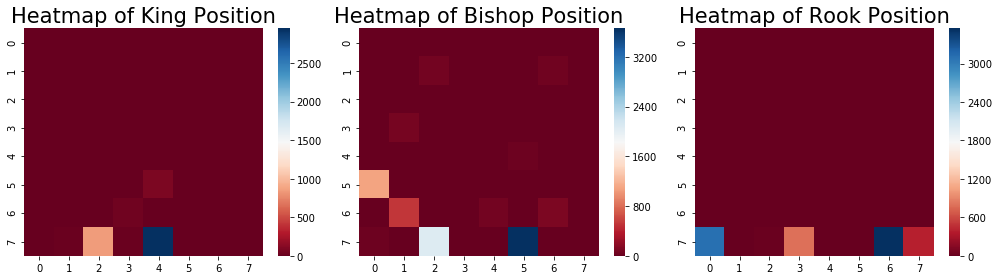}
  \caption{Heatmaps of SentiMATE pieces which show lack of development} 
\end{figure}
\noindent SentiMATE shows a preference for the King Pawn opening, an aggressive and attacking first move, given that it immediately stakes a claim to the center and frees up the queenside bishop. Bobby Fischer (Grand Master) claims that the King's Pawn Game is "Best by test" \cite{fisher}, and stated that "With 1.e4! I win" \cite{fisher2}. This was a common occurrence in our agent's play, and showed its capability of learning given that it was a recurring move in the dataset. It provides control of key central squares and allows for rapid game development. Additionally it can be seen that it has learned to develop its knight specifically king side to deep positions on the board early on. Finally, remarkably, it learns to maintain its queen in an e4 position giving SentiMATE a big presence in the central squares, a region of extreme importance.
\\
\\
On the other hand, Figure 7 shows that the agent does not use its bishop or rooks to their full potential. Rooks are specifically hard to manoeuvre and requires several moves of planning to place them into advantageous positions; this is challenging for SentiMATE, which plans one move ahead. As can be seen in the King's heatmap (in Figure 7), it successfully learns to castle queen side but this does not match well with the highly occurring \textit{b4} pawn seen above. 
\section{Conclusion \& Future Work}
We conclude that this research gives rise to a promising field for move evaluation in Chess engines. It addresses our research hypothesis, given that NLP has successfully been used to train a Chess evaluation function. A dataset of c. 15,000 annotated moves and their respective sentiment is sufficient for the model to outperform a random player and a (restrained) depth-one DeepChess engine. Furthermore, the provided classification models exhibit high accuracy, making the expansion of the dataset both feasible and timely. 
\\
\\
The expansion of the dataset is a key element to the improvement of SentiMATE. DeepChess, employed for comparison purposes, is trained on over one million moves and, whilst we outperformed DeepChess with a depth of one, increasing the size of the provided dataset will allow for the model to have a look-ahead element and plan several moves ahead, allowing for SentiMATE to run on larger depths.
\\
\\
A low-level state representation was used for the model to learn features from scratch. However, it could benefit from having an input with a higher-level of information. This could be done by extracting features which could be deemed beneficial as the input.
\bibliography{main.bib}
\bibliographystyle{aaai.bst}

\end{document}